\documentclass[10pt,twocolumn,letterpaper]{article} 

\usepackage{avss}
\usepackage{times}
\usepackage{epsfig}
\usepackage{graphicx}
\usepackage{amsmath}
\usepackage{amssymb}

\usepackage[ngerman]{babel} % for german oe

\usepackage{url}
\makeatletter
\renewcommand{\paragraph}{%
  \@startsection{paragraph}{4}%
  {\z@}{0.25ex \@plus 1ex \@minus .2ex}{-1em}%
  {\normalfont\normalsize\bfseries}%
}
\makeatother

\usepackage[pagebackref=true,breaklinks=true,letterpaper=true,colorlinks,bookmarks=false]{hyperref}

\avssfinalcopy % *** Uncomment this line for the final submission

\def\avssPaperID{46} % *** Enter the AVSS Paper ID here

\ifavssfinal\fi
\begin{document}

%%%%%%%%% TITLE
\title{Optical Flow Dataset and Benchmark for Visual Crowd Analysis}

\author{Gregory Schr\"oder, Tobias Senst, Erik Bochinski, Thomas Sikora\\
Communication Systems Group\\
Technische Universit\"at Berlin\\
{\tt\small schroeder,senst,bochinski,sikora@nue.tu-berlin.de}
}

\maketitle
% \thispagestyle{empty}

%%%%%%%%% ABSTRACT
\begin{abstract}
The performance of optical flow algorithms greatly depends on the specifics of the content and the application for which it is used. Existing and well established optical flow datasets are limited to rather particular contents from which none is close to crowd behavior analysis; whereas such applications heavily utilize optical flow. We introduce a new optical flow dataset exploiting the possibilities of a recent video engine to generate sequences with ground-truth optical flow for large crowds in different scenarios. We break with the development of the last decade of introducing ever increasing displacements to pose new difficulties. 
Instead we focus on real-world surveillance scenarios where numerous small, partly independent, non rigidly moving objects observed over a long temporal range pose a challenge. By evaluating different optical flow algorithms, we find that results of established datasets can not be transferred to these new challenges. In exhaustive experiments we are able to provide new insight into optical flow for crowd analysis. Finally, the results have been validated on the real-world UCF crowd tracking benchmark while achieving competitive results compared to more sophisticated state-of-the-art crowd tracking approaches. 
%\input{abstract_old}

%focus on real surveillance 
%The introduced crowd optical flow dataset will allow to compare motion estimates on various temporal scales by evaluating two-frame optical flow as well as long-term trajectories. 

%By comparing the performances for the use-case of person tracking in crowds between our synthetic dataset and a real-world tracking benchmark, we could show the portability of the evaluation results to real-world application. 

%By comparing the performances on our dataset with real-world crowd tracking data we could demonstrate that the crowd optical flow dataset is well suited for visual crowd analysis.

%The dataset and the evaluation framework are publicly available.

%Mention long term trajectories (maybe in the introduction):
%Besides motion based approaches for crowd analysis, also trajectory based approaches are utilized. Ground truth for trajectories is also available with hundreds of trajectories per person.

\end{abstract}

\section{Introduction}
Motion estimation based on the principle of optical flow has given rise to a tremendous quantity of work and still is one of the most active research domains in the field of computer vision. 
The history of research on optical flow shows that the accessibility of public benchmarks provided the strongest impetus for significant innovation in the field.
From the first benchmark proposed by Barron \etal \cite{Barron94} in 1994 to more recent e.g. proposed by Butler \etal \cite{Sintel}, the community has benefited greatly from the possibility of a measurable progress in which the limits of technology have been pushed with new and more challenging datasets. 

In visual surveillance, optical flow algorithms have become an important component of crowded scene analysis \cite{JacquesJunior2010,Li2015}.
%
%The application of optical flow allows to measure crowd motion dynamics of hundreds of individuals without the need to detect and track them explicitly, which is an unsolved problem for dense crowds.
%The application of optical flow allows for measuring crowd motion dynamics of hundreds of individuals without the need to detect and track them explicitly, which is an unsolved problem for dense crowds.
The application of optical flow allows crowd motion dynamics of hundreds of individuals to be measured without the need to detect and track them explicitly, which is an unsolved problem for dense crowds.
As a result, optical flow based crowd-motion representations \cite{Mehran2010,Kuhn2012} are a core feature in variety of surveillance applications in e.g. crowd segmentation \cite{Jodoin2013}, crowd behavior analysis \cite{Senst2017} or tracking in crowded scenes \cite{Ali2008}.
However, the impact of the optical flow quality on the crowd analysis has not been sufficiently investigated yet.
In fact, the choice of an appropriate optical flow method for crowd analysis is a challenging issue because the quality of optical flow algorithms can only be stated regarding the specific content and application that is reflected by the recent datasets.
For visual crowd analysis none of the existing optical flow datasets (Middlebury~\cite{Middlebury}, KITTI 2012~\cite{Kitti2012} / 2015~\cite{Kitti2015} MPI-Sintel~\cite{Sintel}) contains suitable content. 

We argue that large crowds show major, non-investigated challenges for optical flow algorithms; 
in particular, the requirements in crowd analysis are: i) precise motion estimation of numerous small, partly independent, self-occluding, non rigidly moving individuals and ii) consistency over a long temporal range.  
In this paper, we propose a new optical flow dataset for visual crowd analysis. 
%
%The dataset comprises over 3200 frames in video sequences with a maximum range up to 450 frames generated with one of the latest video engines.
The dataset comprises over 3200 frames in video sequences ranging up to 450 frames; each generated with one of the latest video engines.
The video engine allows to realistically synthesize thousands of moving individuals simultaneously and acquire ground-truth optical flow fields and person trajectories in different environments simulating five typical crowd analysis scenarios. 
%

%
%Each of the scenarios is rendered with a static and a dynamic camera setup to take modern applications for flying video drones into account which allows to study the impact of the UAV ego-motion.
Each of the scenarios is rendered with a static and a dynamic camera setup to take modern applications for flying video drones into account which allows for studying the impact of the UAV ego-motion.
We will compare the results of state-of-the-art optical flow algorithms for the proposed dataset to their performance on a real-world crowd tracking use-case to show the portability of the benchmark results to real-world crowd surveillance applications.

\section{Related Work}
\label{sec:related_work}
Virtual simulation is a common approach in crowd analysis to study the behavior of complex crowd movements in outdoor and indoor environments.
Especially for high-level events in dense crowds, such as tracing of people flows or the detection of bottlenecks e.g. for infrastructural facility management, virtual simulation has become an indispensable tool. 
Modular frameworks \cite{Narain2009, Curtis2016} allow to design diverse virtual environments with hundreds of moving individuals and generate their exact positions and trajectories.
Due to constant improvements of rendering techniques, synthetic video footage becomes increasingly realistic. 

In contrast, creating comprehensive real-world datasets is time consuming and expensive.
For that reason, nowadays crowd datasets label only a subset of the visible individuals e.g. the UCF crowd tracking dataset~\cite{Ali2008}, or contain only very sparsely annotated crowds \cite{Robicquet2016} or brief video-level based annotations \cite{Dupont2017} describing the crowds rather than the individuals.%, e.g. interacting crowd, static calm, merging flow.

The difficulties to gather annotated real-world data and the high quality of rendering pipelines make the idea of using synthetic data in the field of video surveillance e.g. to evaluate and/or train object-detection, object-tracking or crowd behavior algorithms a promising approach. 
Qureshi and Terzopoulos~\cite{SurveillanceInVirtualReality} proposed a virtual multi-camera system within a train station to evaluate collaborative approaches for tracking of pedestrians. 
It has been shown that detectors trained by virtual data can be transferred and applied to real-world applications.
For example, Mar\'{i}n \etal~\cite{LearningAppearanceInVirtualScenarios} and Hattori \etal~\cite{LearningScene-specificPedestrianDetectors} used synthetic data to train a pedestrian detector without any real-data. 
In~\cite{Bochinski2016} Bochinski \etal utilized the Source game engine to generate synthetic environments with different vehicles, animals and individuals to train a multi-class convolutional neural network for object detection.

%In the field of optical flow, the utilization of synthetic data for benchmarking is a common approach and possibility to challenging datasets from which the community has been benefit greatly.
In the field of optical flow, the community has benefited greatly from synthetic data, where it is commonly used for benchmarking as it allows for creating challenging datasets with sub-pixel accurate ground-truth.
Unfortunately, none of the existing datasets contain crowd analysis related content.
The Middlebury dataset \cite{Middlebury} published in 2007 contains eight short training and eight test sequences from which half of them has been synthetically rendered. 
The main challenge of this dataset is the precise estimation of manifold motion-discontinuities from different large moving or static objects.
The estimated motions are rather small with an average velocity of about 4 and an maximal velocity of 22 pixels.
As the evaluation takes only one optical flow ground-truth field for each sequence into account, it does not allow to check temporal consistency of the motion estimates.

The MPI-Sintel dataset \cite{Sintel} proposed in 2012 is based on the open source 3D animated short film called Sintel. 
The training set consists of 1040 ground-truth optical flow fields from 23 selected sequences. 
The test set contains 564 images spread over 12 sequences.
The average and maximal velocities are 5 and 445 respectively.
The dataset contains a rich set of additional challenges such as long-range motion, illumination changes, specular reflections, motion blur and atmospheric effects.
Taking a closer look reveals that the results of a few extreme challenging sequences with long-range camera or object motions, and strong distortions (e.g. ambush 4) have a dominant impact on the final score. 
Hence, transferring these results to crowd analysis use-cases, where motion of rather small objects is estimated, could be difficult.

Flying Chairs \cite{Dosovitskiy2015} and ChairsSDHom \cite{Ilg2017} are abstract synthetic datasets which are not designed for benchmarking but for training convolutional networks on optical flow. 
%
%An approach between synthetic and natural images has been proposed by Roth and Black \cite{RothBlackRigidDataset}, who used laser scans of real scenes to synthesize optical flow but only for rigid scenes
%
%TODO dieser abschnitt raus?
Liu \etal \cite{Liu2008} developed a semiautomatic tool and published a small dataset, however as Butler \etal state in \cite{Sintel} ``[...]~is not clear that humans are good at segmenting scenes and may inconsistently label regions such as shadows.'' and ``[...]~ground truth flow will always be biased towards a particular algorithm used to compute it.'', which makes the use of this data problematic.

The KITTI~2012 \cite{Kitti2012} and 2015 \cite{Kitti2015} datasets are pure naturalistic benchmarks captured from a car driving through the city of Karlsruhe.
%
%With Middlebury and MPI-Sintel, the KITTI datasets are most common for benchmarking optical flow methods. 
%
The main challenges of these datasets are varying illuminations and long-range motion, i.e. average and maximum velocities are 9 and 549 for KITTI~2012 and 8 and 724 pixels for KITTI~2015. 
Both datasets are specialized for automotive applications and the locomotion of the car has a strong impact to the evaluation results. 

Comparing the results of the four established datasets Middlebury, KITTI~2012/2015 and MPI-Sintel, shows different rankings for the same optical flow methods; not at least because each dataset focuses on a unique subset of issues in the respective field.
We therefore cannot find a clear answer to the question \textit{What is a appropriate optical flow method for crowd analysis?} which raises the need for a dedicated benchmark for this use-case.

%=======
%The concluding observation is that each dataset makes compromises and focuses on a subset of issues in the field and that at the moment there is no consistent  ranking four each of the established datasets: Middlebury, KITTI~2012/2015 and MPI-Sintel.
%%
%With the existing benchmarks we cannot find a clear answer of the question: What is a appropriate optical flow method for crowd analysis?
%>>>>>>> e25ea866957c0a72080bfc5d7bf7fd1233a7800b
%%

%

\begin{figure*}[ht] 
 \small
\setlength{\tabcolsep}{5pt} % Default value: 6pt
\centering
  \begin{tabular}{ccccc} \hline 
        %Sequence & Image Excerpts &  & Average Speed & Angular distribution \\ \hline 
      Sequence  & Sample & Description & Optical flow field & Person trajectories  
			\\ \hline 
        
      % IM01
			\begin{minipage}[c]{0.15\textwidth}
			\scriptsize
			\centering
			IM01 (Static/Dynamic) \\ 371 individuals  \\ 300 frames   
      \end{minipage} &
			\begin{minipage}[c]{0.2\textwidth}
			\includegraphics[width=\textwidth]{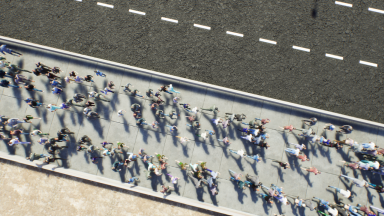}
			\end{minipage} &
			\begin{minipage}[c]{0.15\textwidth}
			\scriptsize
			 %One main stream, few persons walking in opposite direction.
          	Few pedestrians walking against a main crowd flow.
			\end{minipage} &
			\begin{minipage}[c]{0.2\textwidth}
			\includegraphics[width=\textwidth]{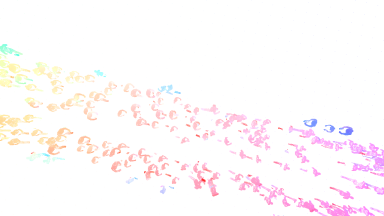}
			\end{minipage} &
			\begin{minipage}[c]{0.2\textwidth}
			\includegraphics[width=\textwidth]{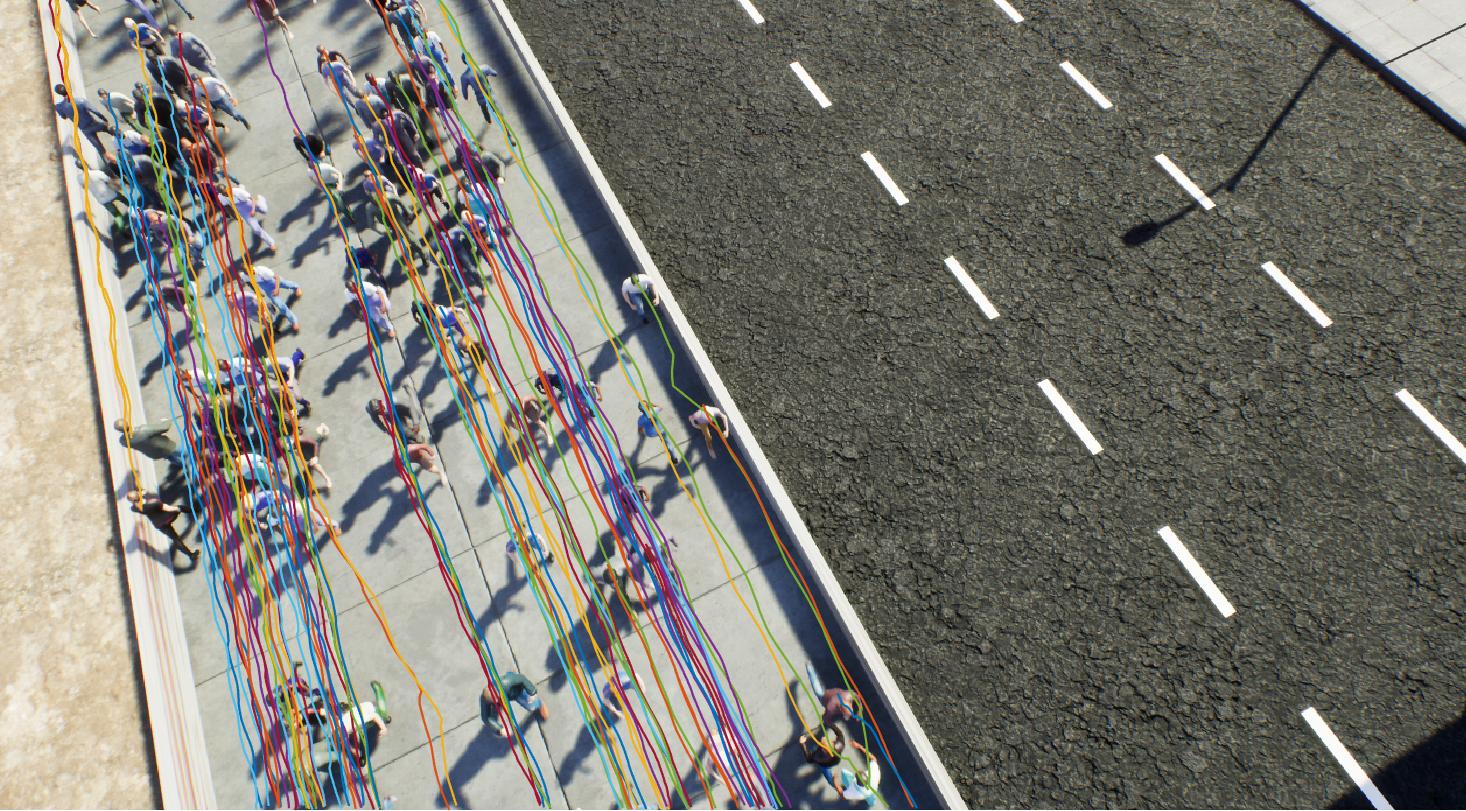}
			\end{minipage}
			%\parbox[c]{7em}{IM01(\_hDyn) \\ 371 individuals  \\ 300 frames \\  } & 
      %\parbox[c]{9em}{\includegraphics[width=1.3in]{figures/ownDataset/IM01_dyn/frame_1118}} & 
      %\parbox[c]{9em}{One main stream, few persons walking in opposite direction.}  & 
      %\parbox[c]{9em}{\includegraphics[width=1.3in]{figures/ownDataset/IM01_dyn/frameGT_1118}} & 
      %\parbox[c]{9em}{\includegraphics[width=1.3in]{figures/ownDataset/IM01/trajectsOverStartimg_cut}} 			
      
      %& 
      %\parbox[c]{5em}{
      %\includegraphics[width=1in]{figures/ownDataset/IM01/trajectsOverStartimg}}  
			\\
			\hline
			\begin{minipage}[c]{0.15\textwidth}
			\scriptsize
			\centering
			IM02 (Static/Dynamic) \\ 631 individuals \\ 300 frames 
      \end{minipage} &
			\begin{minipage}[c]{0.2\textwidth}
			\includegraphics[width=\textwidth]{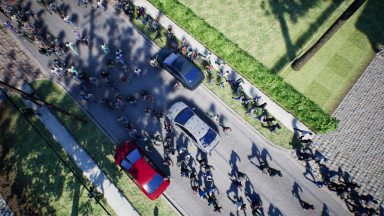}
			\end{minipage} &
			\begin{minipage}[c]{0.15\textwidth}
			\scriptsize
			 %Bottleneck dividing one major stream into three.
            Bottleneck dividing one major flow into three.
			\end{minipage} &
			\begin{minipage}[c]{0.2\textwidth}
			\includegraphics[width=\textwidth]{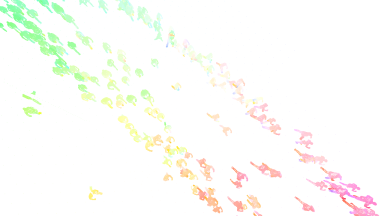}
			\end{minipage} &
			\begin{minipage}[c]{0.2\textwidth}
			\includegraphics[width=\textwidth]{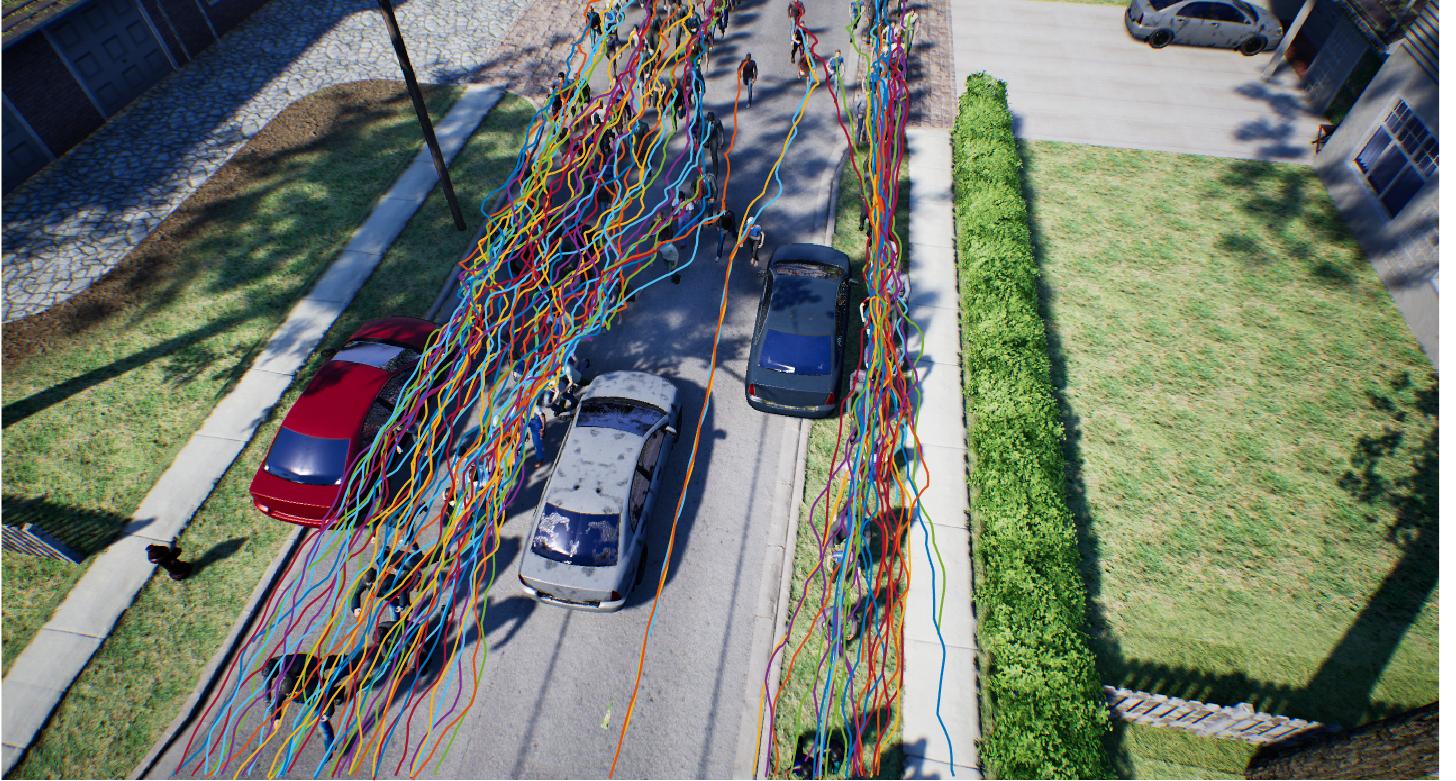}
			\end{minipage}
      % IM02
      %\parbox[c]{7em}{IM02(\_hDyn) \\ 631 individuals \\ 300 frames} & 
      %\parbox[c]{9em}{
      %\includegraphics[width=1.3in]{figures/ownDataset/IM02_dyn/frame_1036}} & 
			%\parbox[c]{9em}{Bottleneck dividing one major stream into three.} 	  &	
      %\parbox[c]{9em}{
      %\includegraphics[width=1.3in]{figures/ownDataset/IM02_dyn/frameGT_1036}} & 
      %\parbox[c]{9em}{
      %\includegraphics[width=1.3in]{figures/ownDataset/IM02/trajectsOverStartimg_cut}}
      \\
      \hline
      \begin{minipage}[c]{0.15\textwidth}
			\scriptsize
			\centering
			IM03 (Static/Dynamic) \\ 878 individuals \\ 250 frames
      \end{minipage} &
			\begin{minipage}[c]{0.2\textwidth}
			\includegraphics[width=\textwidth]{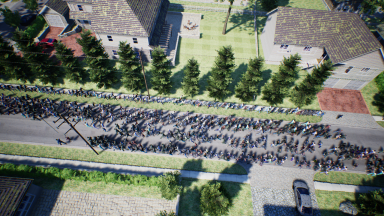}
			\end{minipage} &
			\begin{minipage}[c]{0.15\textwidth}
			\scriptsize
			%Two dense streams walking close past each other.
			Two dense flows walking close past each other.
			\end{minipage} &
			\begin{minipage}[c]{0.2\textwidth}
			\includegraphics[width=\textwidth]{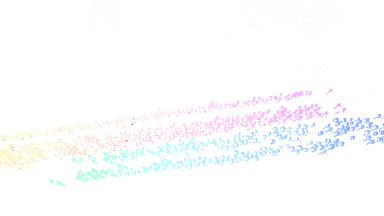}
			\end{minipage} &
			\begin{minipage}[c]{0.2\textwidth}
			\includegraphics[width=\textwidth]{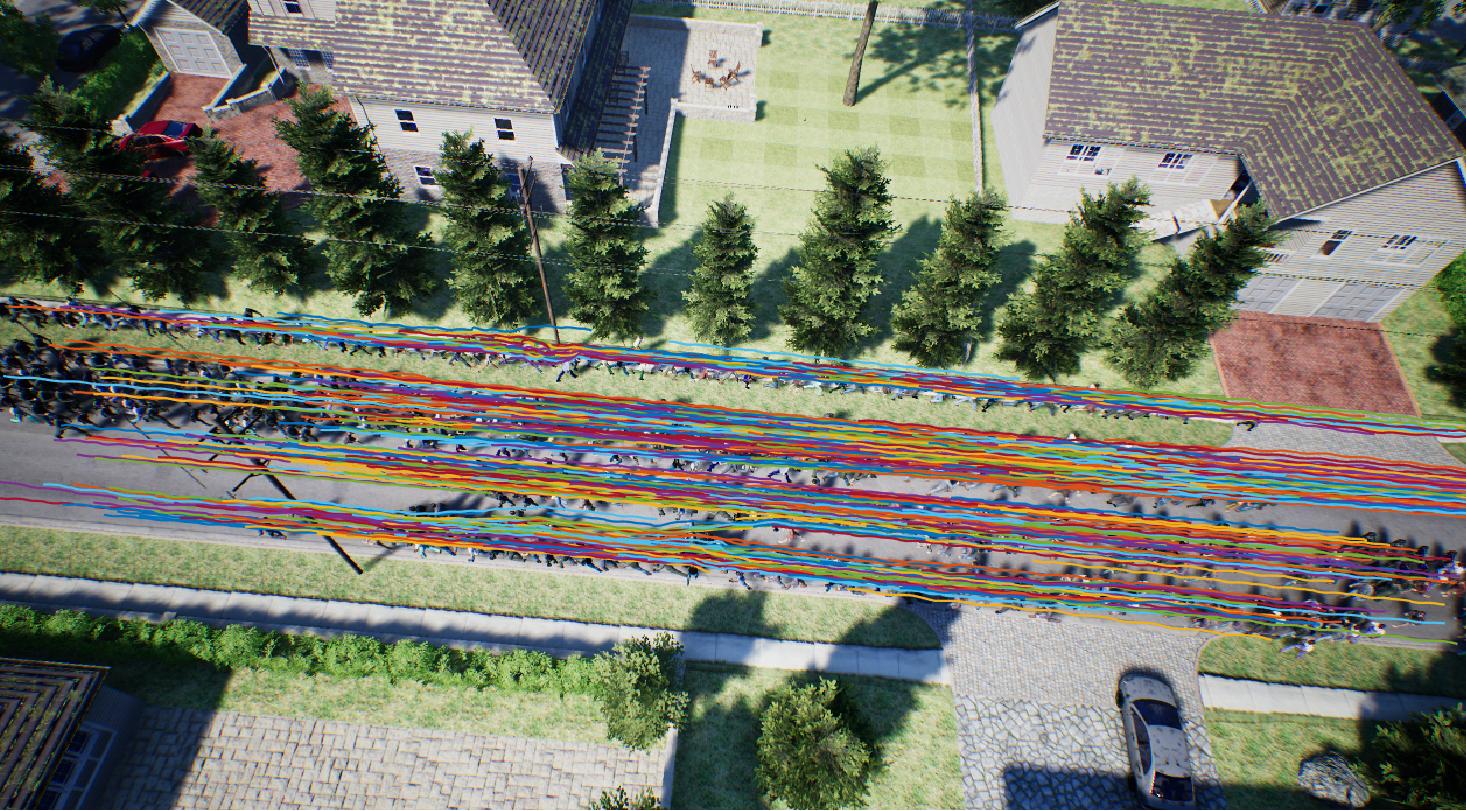}
			\end{minipage}
      % IM03
      %\parbox[c]{7em}{IM03(\_hDyn) \\ 878 individuals \\ 250 frames} & 
      %\parbox[c]{9em}{
      %\includegraphics[width=1.3in]{figures/ownDataset/IM03_dyn/frame_0980}} & 
			%\parbox[c]{9em}{Two dense streams walking close past each other.} &
      %\parbox[c]{9em}{
      %\includegraphics[width=1.3in]{figures/ownDataset/IM03_dyn/frameGT_0980}} & 
      %\parbox[c]{9em}{
      %\includegraphics[width=1.3in]{figures/ownDataset/IM03/trajectsOverStartimg_cut}}
      \\
      \hline
      \begin{minipage}[c]{0.15\textwidth}
			\scriptsize
			\centering
			IM04 (Static/Dynamic) \\ 344 individuals \\ 300 frames
      \end{minipage} &
			\begin{minipage}[c]{0.2\textwidth}
			\includegraphics[width=\textwidth]{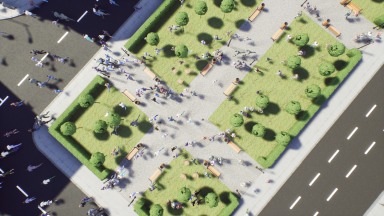}
			\end{minipage} &
			\begin{minipage}[c]{0.15\textwidth}
			\scriptsize
			%Arise of panic and escape.
           	Spread of collective panic and subsequent escape.
			\end{minipage} &
			\begin{minipage}[c]{0.2\textwidth}
			\includegraphics[width=\textwidth]{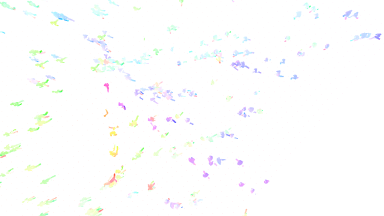}
			\end{minipage} &
			\begin{minipage}[c]{0.2\textwidth}
			\includegraphics[width=\textwidth]{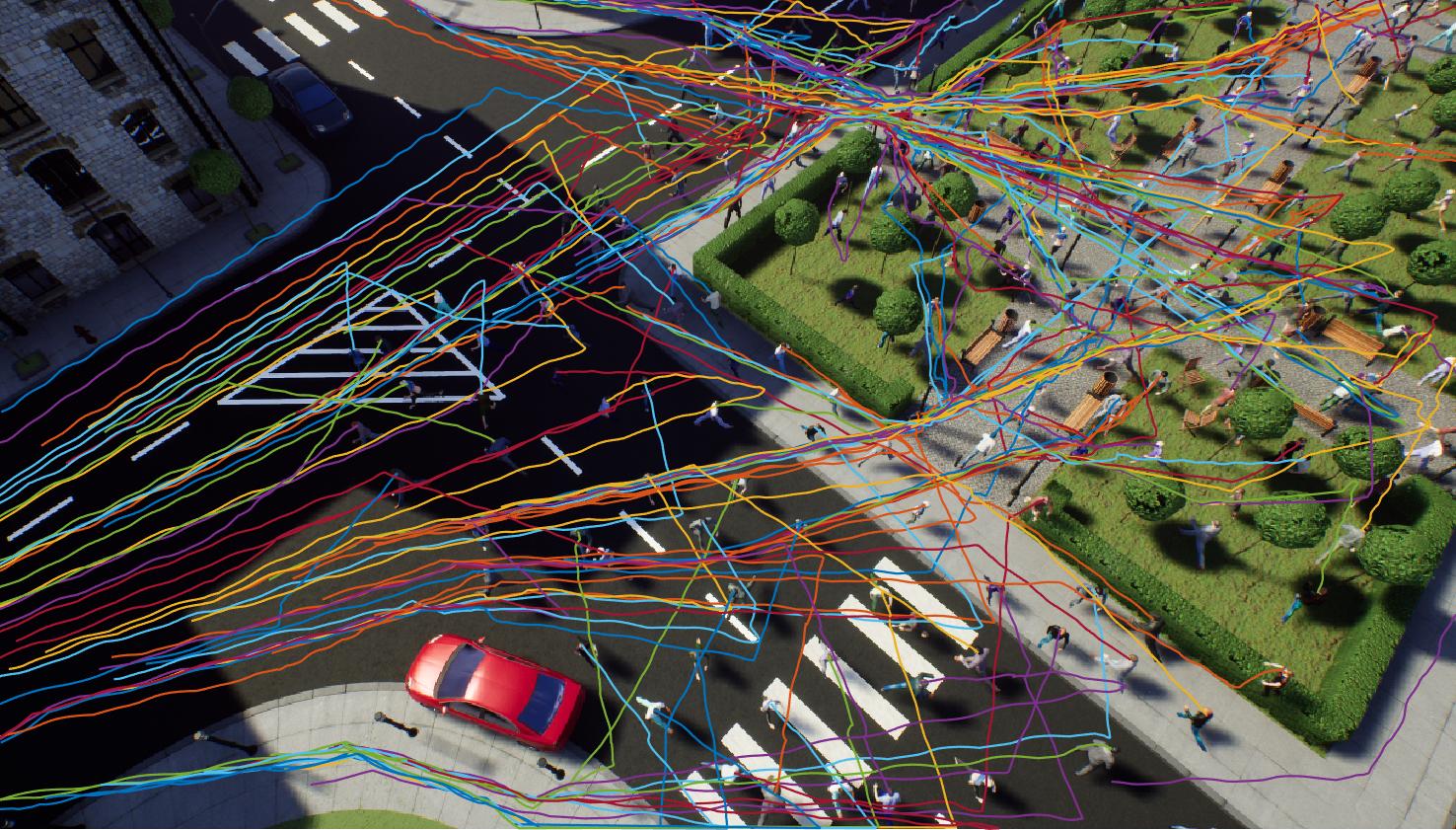}
			\end{minipage}
      % IM04
      %\parbox[c]{7em}{IM04(\_hDyn) \\ 344 individuals \\ 300 frames} & 
      %\parbox[c]{9em}{
      %\includegraphics[width=1.3in]{figures/ownDataset/IM04_dyn/frame_0158}} & 
			%\parbox[c]{9em}{Arise of panic and escape.} 	  		&
      %\parbox[c]{9em}{
      %\includegraphics[width=1.3in]{figures/ownDataset/IM04_dyn/frameGT_0158}} & 
      %\parbox[c]{9em}{
      %\includegraphics[width=1.3in]{figures/ownDataset/IM04/trajectsOverStartimg_cut}}
      \\
      \hline
      \begin{minipage}[c]{0.15\textwidth}
			\scriptsize
			\centering
			IM05 (Static/Dynamic) \\ 1451 individuals \\ 450 frames
      \end{minipage} &
			\begin{minipage}[c]{0.2\textwidth}
			\includegraphics[width=\textwidth]{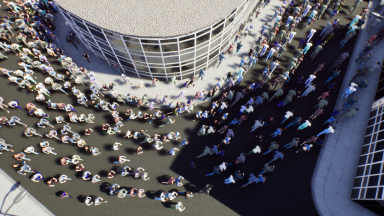}
			\end{minipage} &
			\begin{minipage}[c]{0.15\textwidth}
			\scriptsize
			Marathon sequence. Long temporal tracking.
			\end{minipage} &
			\begin{minipage}[c]{0.2\textwidth}
			\includegraphics[width=\textwidth]{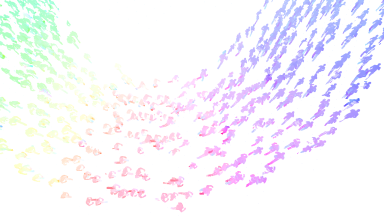}
			\end{minipage} &
			\begin{minipage}[c]{0.2\textwidth}
			\includegraphics[width=\textwidth]{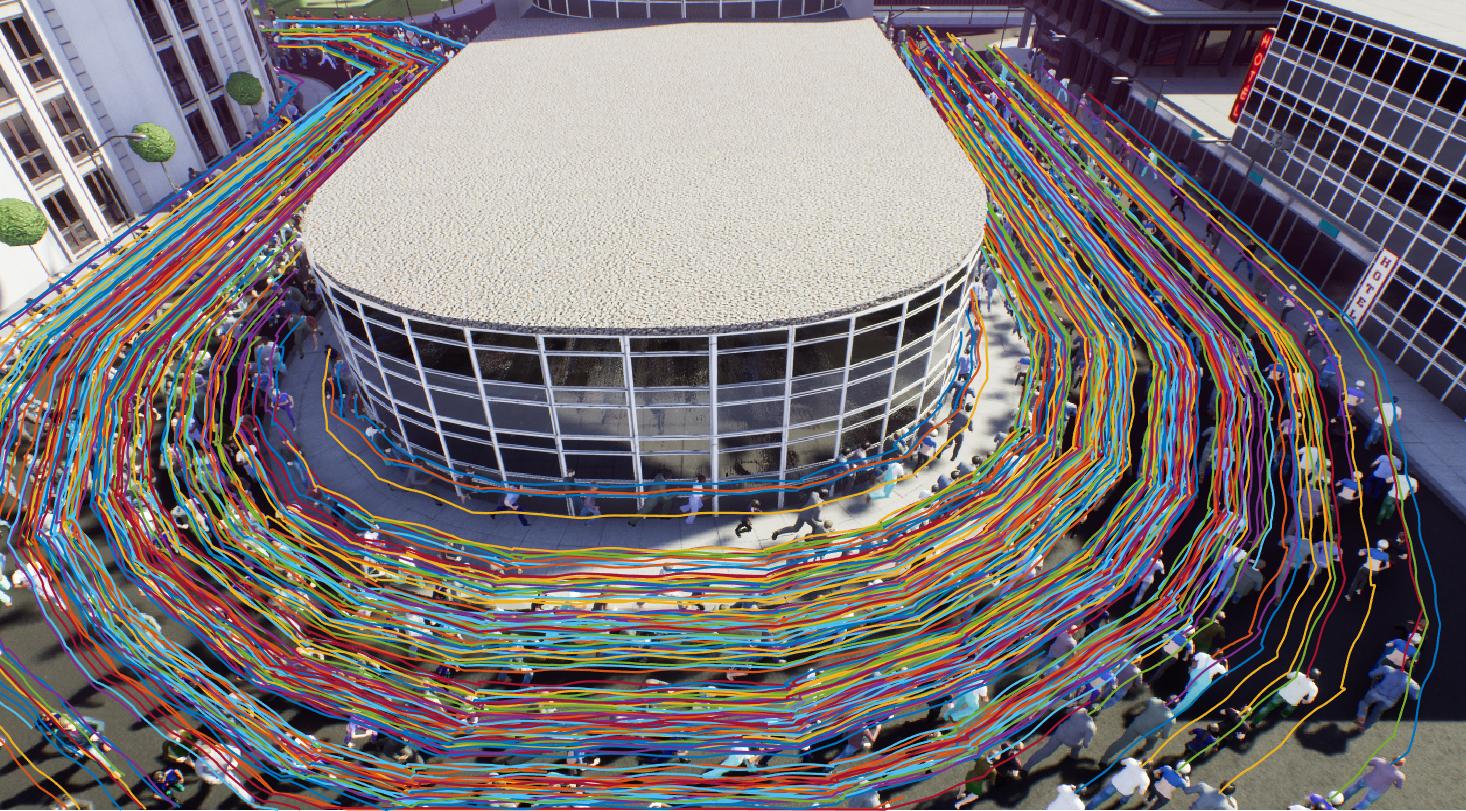}
			\end{minipage}
      % IM05
      %\parbox[c]{7em}{IM05(\_hDyn) \\ 1451 individuals \\ 450 frames} & 
      %\parbox[c]{9em}{
      %\includegraphics[width=1.3in]{figures/ownDataset/IM05_dyn/frame_1621}} & 
			%\parbox[c]{9em}{Marathon sequence. Long temporal tracking.} 	&
      %\parbox[c]{9em}{
      %\includegraphics[width=1.3in]{figures/ownDataset/IM05_dyn/frameGT_1621}} & 
      %\parbox[c]{9em}{
      %\includegraphics[width=1.3in]{figures/ownDataset/IM05/trajectsOverStartimg_cut}}
     \\
      \hline
  \end{tabular}
	%\vspace{0.1cm}
	\caption{Overview of the proposed CrowdFlow dataset with excerpts of the rendered sequences and related ground-truth.} 
\label{fig:CrowdDataset}
\end{figure*}

\section{The Dataset}
In this section we describe our new dataset called CrowdFlow\footnote{available https://github.com/tsenst/CrowdFlow}.
%
%The aim of proposing a new data set is to provide an optical flow benchmark with focus on crowd analysis applications.
It is aimed to provide an optical flow benchmark with focus on crowd analysis applications.
%
%The main purpose of optical flow methods in that field is to estimate the movements of the pedestrians especially in highly crowded scenes.
In that field, the main purpose of optical flow methods is to estimate movements of pedestrians, especially in highly crowded scenes.
%
%The precision of this motion estimation is the prerequisite for subsequent algorithms, such as crowd flow analysis, segmentation or tracking.
A high precision of this motion estimation is an important prerequisite for subsequent algorithms, such as crowd flow analysis, segmentation or tracking.
%To generate scenes in a virtual urban environment, the Unreal Engine is used and allows to simulate hundreds of moving individuals in different environments.
To generate scenes in a virtual urban environment, the Unreal Engine is used which allows to simulate thousands of moving individuals.
The dataset consists of 10 sequences with lengths ranging between 300 and 450 frames.
All sequences were rendered with a frame rate of $25Hz$ and a HD resolution, which is typical for current commercial CCTV surveillance systems.
A comparison to existing optical flow datasets is shown in Tab. \ref{tab:optFlow_stats}.
Besides the increased resolution and number of frames, a major difference to the established datasets is the organization in continuous sequences instead of single frame-pairs (only known from MPI-Sintel), allowing the evaluation of temporal consistencies e.g. in form of trajectories. 

\begin{table}
\centering
\scriptsize
\begin{tabular}{l|c|c|c|c} \hline
Dataset & \# Frames & Rate & Resolution & Year \\ \hline  %  & \# Seq.
Middleburry & 16 & - & $316\times252$ - $640\times480$ & 2007\\ % & 16
MPI-Sintel & 1628 & $24Hz$ & $1024\times436$ & 2012\\  %  & ???
KITTI 2012 & 778 & - & $1242\times375$ & 2012\\  %  & 389
KITTI 2015 & 800 & - & $1242\times375$ & 2015\\  %  & 400
CrowdFlow & 3200 & $25Hz$ & $1280\times720$ & 2018\\ %  & 10
\end{tabular}
\vspace{0.1cm}
\caption{Statistics for existing optical flow benchmarks compared to the proposed CrowdFlow.}
\label{tab:optFlow_stats}
\end{table}

%An overview of the data set can be found in Fig~\ref{fig:CrowdDataset}.
An overview of the sequences, including visualizations of the optical-flow and trajectory ground-truth, is shown in Fig~\ref{fig:CrowdDataset}.
%
%The main design decision for our data set are:
The main design criteria for the dataset are:

\paragraph{Platform:}
%The data set consists of 10 sequences.
%
Each of the 5 unique sequences is rendered twice for different use-case scenarios: one with a static point of view (classic surveillance) and one with a dynamic, airborne point of view (drone/ UAV based surveillance).
This allows to study the impact of a moving camera. 
Further, sudden camera movements ($<50$cm) and angular deviations ($<3^{\circ}$) distort the otherwise smooth camera motion to simulate the typical wind influence on UAVs.
%
%This is due to professional drones having an automatic position regulation and a gimbal embedding for their cameras damping the weather influence.
%
%
\paragraph{Crowd Density:}
None of the recent optical flow benchmarks covers a large amount of differently moving objects.
%
%The CrowdFlow sequences contain between 371 and 1451 independently moving individuals.
%This allows to study the influence between different movements when the crowd is dense or the people occlude each other.
The CrowdFlow sequences contain between 371 and 1451 independently moving individuals.
This allows for the influence between different movements when the crowd is dense or the people occlude each other to be examined.

%The CrowdFlow sequences display between 371 and 1451 moving individuals in varying densities. This allows for the influence of corresponding crowd properties to be examined, e.g. varying individual freedom of motion and people occluding each other.

%
\paragraph{Crowd Movements:}
The scenes cover different kinds of crowd movement:
structured behavior with either a single crowd or two crowds passing each other in different directions as well as fully unstructured movements of the individuals.
%
%Our sequences are 250, 300 and 450 frames long with 249, 299 and 449 ground-truth flow fields and thus 5 -- 9 times longer compared to MPI-Sintel (50 frames).
%
%This allows to study the temporal consistency of the optical flow fields their ability to track each person in the crowd over a longer temporal range.
%
\paragraph{Temporal Consistency:}
%The temporal consistency of motion estimates has not been investigated in recent optical flow benchmarks yet.
%
Maintaining consistent flow fields over a long temporal range is a new challenge in the proposed dataset which is not covered by recent optical flow benchmarks yet.
It allows for analyzing optical flow fields as time-depended vector fields, thus being able to measure related errors such as drifting. 
%The motivation is that inconsistent flow fields lead to errors in field lines, such as pixel trajectories or streak-lines flow, and thus in the crowd motion representation.
%
%We provide a set of dense ground-truth trajectories on crowded regions and evaluate the corresponding trajectories derived from optical flow fields generated using various state-of-the-art methods to analyze the consistency of the quality of the flow fields over time.

%To analyze the consistency of the quality of optical flow fields over time, we provide a set of dense ground-truth trajectories of the crowded regions.
%Corresponding trajectories can be derived directly from the flow fields to be tested and evaluated.

%
%Temporal consistency refers especially to use-cases for crowd segmentation.
%
%\paragraph{Tracking in Crowds:}
%The use-case of tracking in crowds was chosen to test the portability of the benchmark results.
%
%In order to do so, one ground-truth trajectory per person, located at head (in accordance to \cite{Haroon2014}) is extracted.
%The respective trajectories derived from the test field can be evaluated using common tracking metrics.
%Therefore, we extract specific ground-truth trajectories located at the pedestrian heads and compare to corresponding generated trajectories for the test fields.
%
%In extend to analyzing optical flow methods, this dataset can be used to evaluate arbitrary crowd tracking approaches with the same evaluation protocol used in e.g. the UCFCrowd tracking in crowds data set.
%
\paragraph{Portability:}
Being able to transfer the benchmark results to real-world use-cases is a main criteria for synthetic datasets.
In our experiments, we therefore evaluate and compare the performances of several state-of-the-art optical flow methods with respect to the crowd tracking accuracy on the proposed synthetic and the real-world UCF crowd tracking datasets~\cite{Ali2008}.
%
%To create similar conditions we designed the sequences IM01, IM03 and IM05 resembling the respective sequences Seq1, Seq3 and Seq5 of the UCF crowd tracking dataset. 
To create similar conditions we designed the sequences IM01 and IM05 resembling the respective sequences Seq1 and Seq5 of the UCF crowd tracking dataset. 

\begin{figure}
%trim=left bottom right top
%width=0.52\textwidth
\scriptsize
\setlength{\tabcolsep}{3pt} % Default value: 6pt
\begin{tabular}{ccc}
\includegraphics[trim=6.4cm 10cm 6.5cm 10.1cm,clip,width=0.155\textwidth]{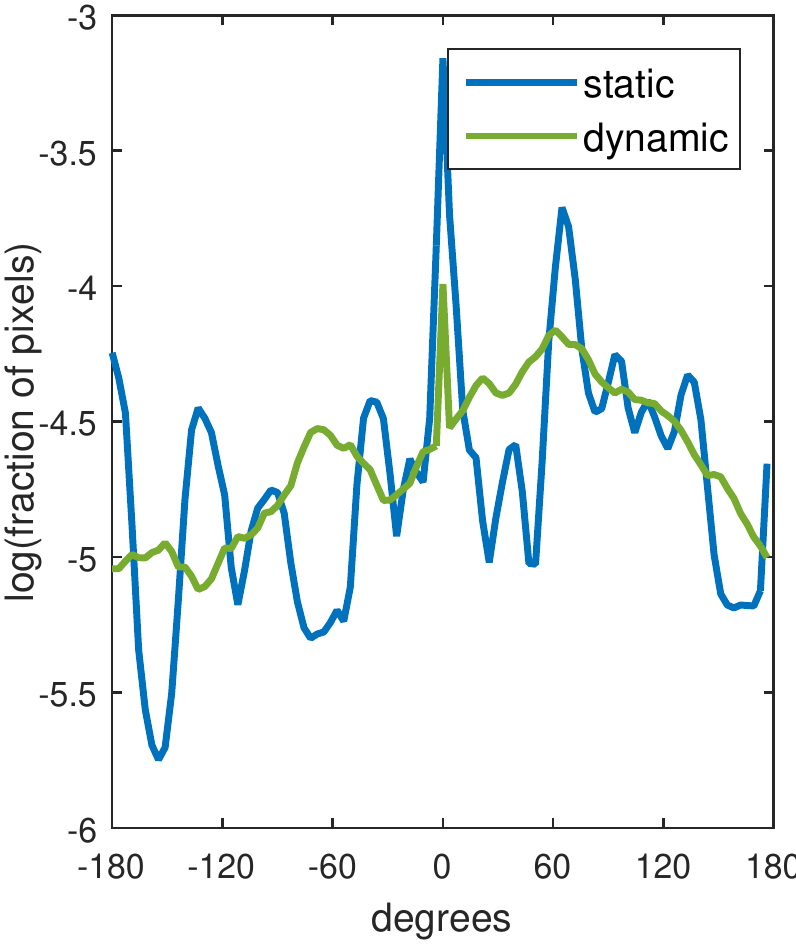} &
\includegraphics[trim=6.4cm 10cm 6.5cm 10.1cm,clip,width=0.155\textwidth]{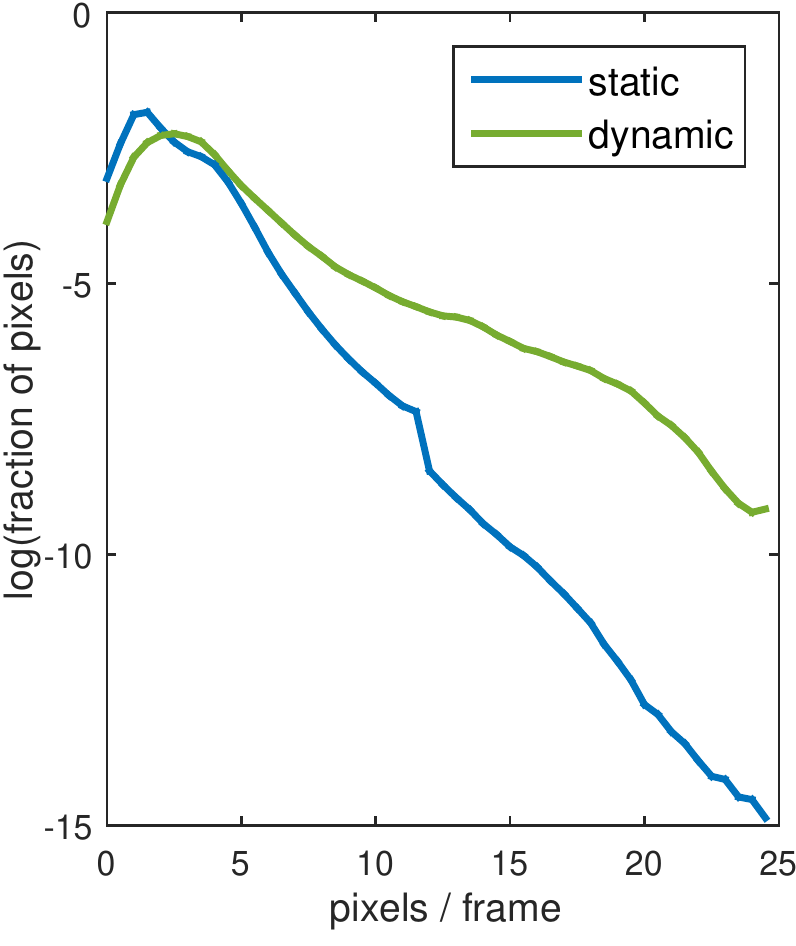} &
\includegraphics[trim=6.5cm 10cm 6.5cm 10.2cm,clip,width=0.153\textwidth]{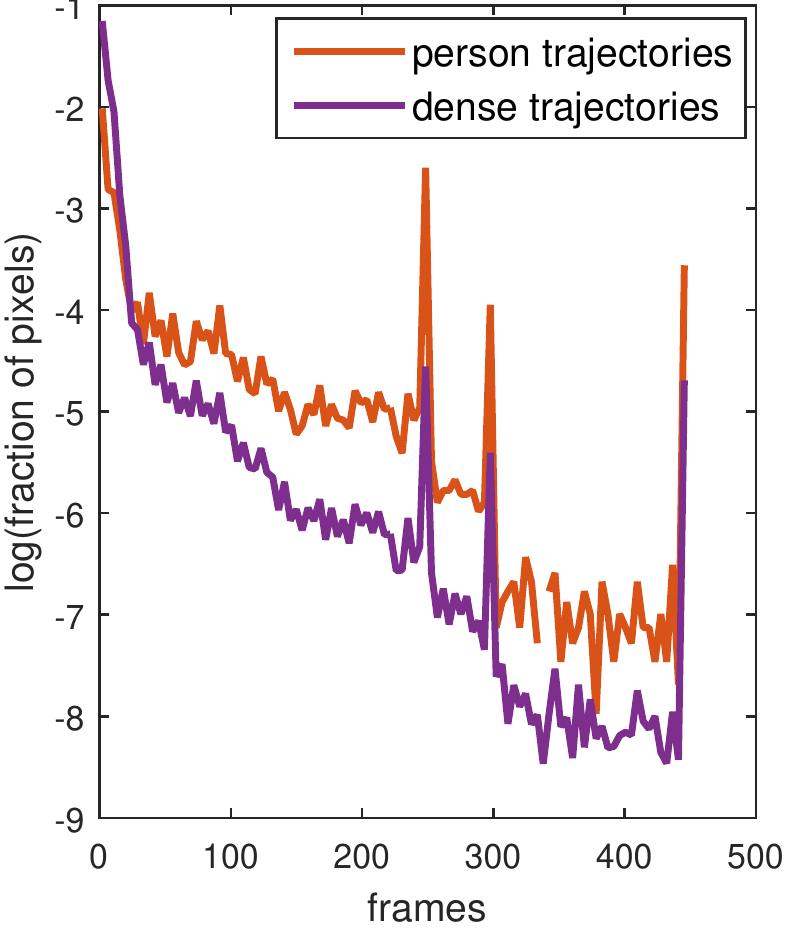} \\
a) flow direction &  b) flow speed & c) trajectory length \\
\end{tabular}
\caption{Statistics of the ground-truth optical flow fields (a-b) and ground-truth trajectories (c).}
\label{fig:statistics}
\end{figure}
Two types of ground-truth data are provided: optical flow fields and trajectories.
Examples can be found in Fig.~\ref{fig:CrowdDataset}.

\textit{Optical Flow:}
The optical flow ground-truth is divided into two categories: foreground and background.
For the foreground, the dense flow for all pixels associated with the pedestrians is provided.
In addition, the background motion is supplied on a sparse grid-like structure as it may also be of interest e.g. for global motion estimation applications.
%kann dieser satz weg?
%Binary masks are provided to differentiate both types of ground truth.
%
%A sparse flow field provides the ground-truth optical flow for the humans motion.
%
%It is derived from a high-resolution human model vertex-mesh and provides dense flow for each pedestrian.
%
%The background motion is recorded on a sparse grid-like structure.
%
%As we focus on crowd motion we provide a binary mask to measure foreground/crowd and background motion.
%

\textit{Trajectories:}
To provide a deeper insight into the temporal consistency of the optical flow fields, the ground-truth contains dense and sparse trajectories for each individual.
The dense trajectories cover almost all visible pixels of the individuals until they get occluded by other persons, objects or body-parts.
This trajectory set allows to study the temporal consistency of the estimated motions per individual over several frames.
The person trajectories are located at the head, similar to \cite{Haroon2014}, thus allowing comparable evaluations for tracking in crowds.
%
%Beside the justification of the data set, this trajectory set allows to infer a more person oriented perspective of tracking related properties such as drifting.

The statistics of both ground-truth data is given in Fig.~\ref{fig:statistics}.

\section{Evaluation Metrics}
To assess the quality of the optical flow we propose to use two types of metrics: \textit{i) common optical flow metrics}, i.e. average endpoint error (EPE) and percentage of erroneous pixel (RX) and \textit{ii) long-term motion metrics} based on trajectories.
Additionally, the run-time is a critical measure to assess the usability for real-time applications.

\paragraph{Optical Flow Metrics:} For each sequence, the EPE and R2 values will be reported.
While the EPE maps over the total error range, the R2 indicates the percentage of pixels with an end-point error larger than two. 
With R2, we set a tolerance error threshold to half of the average body size which is four pixels in our data set. 
To bundle the sequence results for the whole dataset the average of the sequence EPE and R2 are computed.

\paragraph{Long-term Motion Metrics:} 
%For the optical flow field to test, trajectories are seeded at the starting points of the  dense or person ground-truth trajectories advected by optical flow fields. 
To evaluate the optical flow fields, trajectories are seeded at the starting points of the  dense or person ground-truth trajectories and advected by these. 
While the propagated trajectory points are in the sub-pixel domain and the motion vectors are defined on the discrete pixel grid, we found a bilinear interpolation to be sufficiently accurate to reconstruct the corresponding motion vector.
% 
%The trajectory approach allows to evaluate the optical flow fields time-depending.
The trajectory approach allows for a time-depending evaluation of the optical flow fields.
We follow the tracking accuracy proposed in \cite{Haroon2014} for quantitative evaluations.
This metric measures accumulative motion errors and disruptions from temporal inconsistencies of the flow fields.
The tracking accuracy reports the percentage of tracked points from all trajectories that lie within a certain distance to the corresponding ground-truth points.
As in \cite{Dehghan2018} we will use an error threshold of 15 for the qualitative comparison.

\section{Experimental Results}
We evaluated six state-of-the-art optical flow algorithms: RIC~\cite{Hu2017}, CPM~\cite{Hu2016} and FlowFields~\cite{Bailer2015} which are highly accurate approaches and currently ranked in the uppermost quarter of the MPI-Sintel benchmark, DeepFlow~\cite{Weinzaepfel2013}, and DIS~\cite{Kroeger2016} and RLOF~\cite{Geistert2016} which are the top run-time efficient approaches.
Each implementation is online available and supplies a set of baseline configurations. 
In our experiments, we only report results of those configurations which achieved the best performance for dense trajectories of the proposed dataset.
%
%For DIS and RLOF we report two configurations: DIS$^2$ (parameter setup 2, see \cite{Kroeger2016}) and RLOF$^{6}$ (grid size 6, see \cite{Geistert2016}) with run time optimized parameters, and DIS$^4$ and RLOF$^{10}$  with precision optimized parameters.
For DIS and RLOF we report two configurations: DIS$^2$ (parameter setup 2, see \cite{Kroeger2016}) and RLOF$^{10}$ (grid size 10, see \cite{Geistert2016}) with run time optimized parameters, and DIS$^4$ and RLOF$^{6}$ with precision optimized parameters.

\begin{table*}
\scriptsize
\setlength{\tabcolsep}{3pt}

\centering
\begin{tabular}{l|crcr|crcr|crcrcr|r} 
\hline
 \multicolumn{1}{c|}{} & \multicolumn{2}{|c}{FG (Static) } &\multicolumn{2}{c|} { BG (Static)} & \multicolumn{2}{|c}{FG (Dynamic)} & \multicolumn{2}{c|}{ BG (Dynamic)} &\multicolumn{2}{c}{FG($\varnothing$)}&\multicolumn{2}{c}{BG ($\varnothing$)} & \multicolumn{2}{c|}{$\varnothing$} & \\ 
 
 \multicolumn{1}{c|}{}& EPE & R2[\%] & EPE & R2[\%]& EPE & R2[\%]& EPE & R2[\%]& EPE & R2[\%]& EPE & R2[\%]& EPE & R2[\%] & t[sec]\\ 
\hline

FlowFields & 0.756 & 8.27 & \textbf{0.213} & \textbf{2.79} & 1.069 & 14.92 & \textbf{2.571}
& \textbf{51.42} & 0.913 & 11.595 & \textbf{1.392} & \textbf{27.10} & 0.915 & 11.74 & 43.53\\ 

RIC        &  0.859 & 8.64 & 0.243 & 3.31 & 1.166 & 15.69 & 2.623 &
53.58 & 1.013 & 12.164 & 1.433 & 28.45 & 1.015 & 12.32 & 8.30 \\  

CPM        & 0.701 & 7.09 & 0.247 & 3.63 & 1.026 & 13.94 & 2.585 & 
51.78 & 0.864 & 10.517 & 1.416 & 27.71 & 0.868 & 10.69 & 14.74\\ 

DeepFlow   & 0.629 & 6.19 & 0.237 & 3.67 & 1.005 & 13.95 & 2.594
& 51.67 & 0.817 & 10.069 & 1.416 & 27.67 & 0.822 & 10.25 & 39.63\\ 

RLOF$^6$   & 0.753 & 8.61 & 0.315 & 5.00 & 1.088 & 15.61 & 2.655
& 53.47 & 0.921 & 12.112 & 1.485 & 29.23 & 0.924 & 12.27 & 1.49\\ 

RLOF$^{10}$& 0.772 & 8.80 & 0.324 & 5.10 & 1.104 & 15.80 & 2.658 
& 53.60 & 0.938 & 12.303 & 1.491 & 29.35 & 0.941 & 12.46 & 0.80\\ 

DIS$^4$    & \textbf{0.627} & \textbf{5.72} & 0.356 & 5.85 & \textbf{0.928} & \textbf{11.86} & 2.665 
& 53.67 & \textbf{0.777} &  \textbf{8.790} & 1.511 & 29.76 & \textbf{0.784} & \textbf{9.01}  & 1.70\\ 

DIS$^2$    & 1.441 & 20.40 & 0.528 & 8.24 & 1.726 & 27.41 & 3.001 
& 64.01 & 1.583 & 23.903 & 1.765 & 36.13 & 1.579 & 23.92 & \textbf{0.28}\\ 
\hline
\end{tabular}
\vspace{0.1cm}
\caption{Evaluation results on the proposed CrowdFlow data set with \textbf{common optical flow metrics}. Dynamic comprised sequences with and static without camera motion, BG - background motion vectors and FG - motion vectors located at persons of the crowd. $t$ denotes the average processing time on a Intel i9-7980XE CPU @ 2.60 GHz in multi-threading mode.} %for sequence IM01
\label{tab:Results_two_frame_flow}
\end{table*}

Table~\ref{tab:Results_two_frame_flow} shows the comparative results for EPE, R2 and the run-time.
In summary, each approach tends to achieve accurate results, except for DIS$^2$ and with an EPE above 1.5 pixel. 
Overall, the most precise method is DIS$^4$.
It is worth to note that the highly accurate approaches are no more precise than the fast processing ones when estimating crowd movement.
In the presence of additional camera motion the precision of each approach deteriorates significantly. 
Even for static scenes the background contains motion estimation errors, whereby the majority is caused by too homogeneous textures of the streets.
Here, the background motion is biased by neighboring crowd motion vectors and smoothing effects of regularization terms or interpolation errors in case of CPM, RIC and FlowFields.

\begin{table*}
%\tiny
\scriptsize
\setlength{\tabcolsep}{2.4pt}

\centering
\begin{tabular}{l|cc|cc|cc|cc|cc|c|cc|cc|cc|cc|cc|c} 
	\hline
 & \multicolumn{11}{|c|}{Dense Trajectories} & \multicolumn{11}{|c}{Person Trajectories} \\
\cline{2-23}

 & \multicolumn{2}{c|}{IM01 (Dyn)} & \multicolumn{2}{c|}{IM02 (Dyn)} 
& \multicolumn{2}{c|}{IM03 (Dyn)} & \multicolumn{2}{c|}{IM04 (Dyn)} & \multicolumn{2}{c|}{IM05 (Dyn)} &   $\varnothing$ 
 & \multicolumn{2}{c|}{IM01 (Dyn)} & \multicolumn{2}{c|}{IM02 (Dyn)} 
& \multicolumn{2}{c|}{IM03 (Dyn)} & \multicolumn{2}{c|}{IM04 (Dyn)} & \multicolumn{2}{c|}{IM05  (Dyn)} &   $\varnothing$\\ 
 \hline 
FlowFields & 70.63 & 61.79 & 56.69 & 45.93 & 71.46 & 68.35 & 42.27 & 37.63 & 65.15 & 59.61 & 57.95 & 
77.94 & 62.68 & 52.35 & 38.22 & 66.76 & 63.17 & 30.09 & 25.24 & 65.67 & 68.20 & 55.03 \\ 

RIC 	   & 74.39 & 69.41 & 58.72 & 50.33 & 54.18 & 73.80 & 44.21 & 39.52 & 60.23 & 60.28 & 58.51 & 
87.88 & 80.87 & 56.56 & 48.14 & 43.49 & 70.98 & 32.48 & 27.81 & 57.47 & 68.56 & 57.42\\

CPM 	   &  73.41 & 65.16 & 58.31 & 47.57 & 74.41 & 71.13 & 46.23 & 41.15 & 67.97 & 61.68 & 60.70 & 
82.17 & 68.82 & 54.56 & 40.99 & 70.37 & 66.69 & 35.98 & 30.00 & 69.64 & 71.58 & 59.08 \\ 

DeepFlow   & \textbf{83.84} & \textbf{81.90} & \textbf{63.33} & 55.52 & 83.38 & 80.87 & \textbf{57.08} & \textbf{56.65} & 71.25 & 64.67 & \textbf{69.85} & 
\textbf{99.19} & \textbf{95.32} & \textbf{68.60} & 63.04 & 83.18 & 81.20 & \textbf{53.82} & \textbf{52.22} & \textbf{76.32} & 79.15 & \textbf{75.20} \\ 

RLOF$^6$   & 82.80 & 78.31 & 63.16 & \textbf{57.68} & \textbf{87.46} & \textbf{86.76} & 50.56 & 50.53 & 69.86 & 68.73 & 69.59 & 
97.70 & 92.37 & 66.70 & \textbf{65.08} & \textbf{88.73} & \textbf{90.22} & 43.56 & 46.47 & 72.60 & 80.12 & 74.36\\ 

RLOF$^{10}$& 80.14 & 73.95 & 62.05 & 55.54 & 85.44 & 84.39 & 48.80 & 47.84 & 67.53 & 67.41 & 67.31 & 
96.00 & 85.02 & 63.08 & 59.77 & 85.97 & 86.69 & 39.41 & 40.48 & 69.09 & 78.70 & 70.42\\ 

DIS$^4$    & 80.44 & 76.19 & 64.11 & 56.99 & 82.89 & 82.24 & 53.91 & 52.75 & \textbf{72.11} & \textbf{70.71} & 69.23  & 
92.22 & 85.98 & 63.97 & 56.35 & 81.59 & 81.61 & 44.58 & 42.64 & 74.95 & \textbf{82.09} & 70.60 \\ 

DIS$^2$    & 47.55 & 33.03 & 36.52 & 25.32 & 22.59 & 19.76 & 26.79 & 20.89 & 27.63 & 27.91 & 28.80 & 
40.81 & 22.39 & 22.86 & 15.37 & 9.05 & 6.72 & 13.63 & 9.72 & 17.86 & 18.10 & 17.65 \\

\hline
\end{tabular}
\vspace{0.01cm}
\caption{Evaluation results on CrowdFlow data set with long-term motion metric. The \textbf{tracking accuracy} in percentage for the threshold set to 15 pixels. Higher values denote more accurate results.}
\label{tab:Results_Long_term}
\end{table*}

Table~\ref{tab:Results_Long_term} shows the results with respect to the tracking accuracy.
While the flow fields accuracy for this dataset is on a frame-based level (EPE and R2) already quite high, the accuracy of the time-depended perspective of the tracking accuracy poses a significant challenge for the existing methods.
None of the evaluated methods achieved an accuracy above 70\% for the dense trajectories and 76\% for the person trajectories. 
In contrast to the frame-based results, DeepFlow is on average the most accurate approach, with RLOF${^6}$ and DIS$^{4}$ achieving similar performances for the dense trajectories.
An interesting observation is that RLOF${^6}$ is very accurate on the long-term basis, while it achieves only moderate results for common optical flow metrics. 
All algorithms perform worse on dynamic sequences compared to the static ones.

The evaluation results of the flow methods for the real-world UCF crowd tracking benchmark is depicted in Table~\ref{tab:UCF_Crowd_Tracking}.
In addition, we report tracking performances of the state-of-the-art in that area.
Although the trajectories are only computed by simple bilinear interpolation, the optical flow methods achieve competitive results.
It shows that methods considered to be highly accurate such as FlowFields, RIC and CPM also behave less accurate than DeepFlow, RLOF and DIS.
%
%Meanwhile, the ranking for the UCF crowd tracking is consistent to the proposed for CrowdFlow. 
Meanwhile, the ranking for the UCF crowd tracking is consistent to the proposed CrowdFlow dataset and also its quantitative results are similar. 
Note that due to the higher resolution of the CrowdFlow sequences the tracking accuracy threshold of 15 is a stricter measurement compared to the lower resolution ($720\times480$ or less) of the UCF crowd tracking benchmark.
With this prove of concept, we show that our synthetic dataset is better suitable to assess optical flow algorithms for crowd analysis than existing optical flow benchmarks.
%

%\begin{table*}
%
%\small
%%\footnotesize % even smaller
%\centering % centers the whole table
%\scriptsize
%\begin{tabular}{lllllllllllll} \hline
%		 	& Seq1 & Seq2 & Seq3 & Seq4 & Seq5 & Seq6 & Seq7 & Seq8 & Seq9 & Seq10 & Seq11 &  $\varnothing$ 	\\ \hline 
%
%DeepFlow 	  &62 &100 &89 &97 &62 &84 &65 &91 &34 &82 & 47 & 73.91 \\ 
%CPM         & 50.34 & 99.59 & 86.04 & 95.71 & 40.01 & 82.88 & 61.73 & 87.02 & 23.80 & 39.41 & 23.73 & 62.75\\ 
%RIC         & 38.94 & 99.63 & 91.68 & 94.22 & 35.16 & 85.06 & 63.68 & 87.86 & 22.72 & 44.51 & 41.40 & 64.08\\ 
%FlowFields  & 50.34 & 99.59 & 86.02 & 95.88 & 40.00 & 82.87 & 61.71 & 87.01 & 23.81 & 39.40 & 23.69 & 62.76\\ 
%RLOF$^{10}$ & 62.59 & 99.59 & 90.94 & 95.77 & 57.18 & 87.80 & 66.90 & 88.22 & 32.88 & 62.58 & 74.03 & 74.41\\
%RLOF$^6$    & 63.80 & 99.66 & 91.17 & 96.07 & 60.39 & 88.51 & 67.02 & 89.55 & 35.63 & 66.03 & 77.10 & 75.90\\
%\hline
%BQP & 86 & 99 & 96 & 97 & 78 & 96 & 67 & 90 & 78 & 92 & 91 & 88.18\\
%NMC & 80 & 100 & 92 & 94 & 77 & 94 & 67 & 92 & 63 & - & - & ?\\
%Floorfields & 74 & 99 & 83 & 88 & 66 & 90 & 68 & 93 & 47 & - & - & ? \\
%\end{tabular}
%\caption{
%Tracking accuracy on CRCV dataset for pixel threshold T = 15 with flowScale factor = 1 in percent.} %The bold numbers indicate the best performing optical flow algorithms.}
%\end{table*}

\begin{table}

\small
%\footnotesize % even smaller
\setlength{\tabcolsep}{2.4pt}
\centering % centers the whole table
\scriptsize
\begin{tabular}{lcccccccccc} \hline
		 	& Seq1 & Seq2 & Seq3 & Seq4 & Seq5 & Seq6 & Seq7 & Seq8 & Seq9 & $\varnothing$ 	\\ \hline 
FlowFields  & 50 & \textbf{100} & 86 & \textbf{96} & 40 & 83 & 62 & 87 & 24 &  69.78\\ 
RIC         & 39 & \textbf{100} & \textbf{92} & 94 & 35 & 85 & 64 & 88 & 23 & 68.89\\ 
CPM         & 50 & \textbf{100} & 86 & \textbf{96} & 40 & 83 & 62 & 87 & 24 & 67.33\\ 
DeepFlow 	  %& 62 & \textbf{100} & 89 & \textbf{97} & \textbf{62} & 84 & 65 & \textbf{91} & 34 & 76.00 \\ 
						& 60 & \textbf{100} & 88 & \textbf{96} & 59 & 84 & 65 & 89 & 33 & 71.56\\
RLOF$^6$    & 64 & \textbf{100} & 91 & \textbf{96} & \textbf{60} & \textbf{89} & \textbf{67} & \textbf{90} & \textbf{36} &  \textbf{77.00}\\
RLOF$^{10}$ & 63 & \textbf{100} & 91 & \textbf{96} & 57 & 88 & \textbf{67} & 88 & 33 & 75.89\\
DIS$^4$     & \textbf{71} & \textbf{100} & \textbf{92} & 96 & 46 & 88 & 63 & 89 & 31 & 75.11\\
DIS$^2$     & 54 & 66  & 86 & 83 & 16 & 80 & 35 & 64 & 19 &  55.89 \\
\hline
BQP         & 86 & 99  & 96 & 97 & 78 & 96 & 67 & 90 & 78 & 87.44\\
NMC         & 80 & 100 & 92 & 94 & 77 & 94 & 67 & 92 & 63 & 84.33\\
Floorfields & 74 & 99  & 83 & 88 & 66 & 90 & 68 & 93 & 47 & 78.67 \\
\hline
\end{tabular}
\vspace{0.01cm}
\caption{Evaluation results on UCF crowd tracking dataset \cite{Ali2008} based on tracking accuracy with the threshold set to 15. 
Bottom rows show state-of-the-art tracking methods for this dataset: BQP~\cite{Dehghan2018}, NMC~\cite{Haroon2014} and Floorfields~\cite{Ali2008}.} 
\label{tab:UCF_Crowd_Tracking}
\end{table}

%RLOFMode RIC 1  Runtime  0.611605223206 9.88606828192

\section{Conclusion}
In this paper, we presented a novel optical flow benchmark targeting crowd analysis applications. 
In contrast to previous benchmarks, our sequences contain up to 1451 partly independent moving individuals which poses a new challenge.
To cover classic and modern UAV based surveillance scenarios, we rendered each sequence with static and dynamic camera views. 
This gives us the unique opportunity to study the impact of non-stationary camera setups.
We introduced a trajectory based long-term metric, which is new to optical flow benchmarks, to capture time-dependent motion estimation errors like drifting.
In our experiments, we showed that these metrics are more discriminative than the common optical flow metrics such as EPE when it comes to crowd related analysis like tracking.
%
%We showed that the ranking of state-of-the-art flow algorithms on our CrowdFlow benchmark is significantly different from on existing benchmarks.
We showed that the ranking of state-of-the-art flow algorithms on our CrowdFlow benchmark differs significantly from existing benchmarks.
In experiments on the real-world UCF crowd tracking dataset, we confirmed our ranking indicating the usefulness of our benchmark approach for such applications.

%\bigskip \noindent
\section{Acknowledgment}
%\paragraph{Acknowledgment:}
The research leading to these results has received funding
from  BMBF-VIP+ under grant agreement number 03VP01940
(SiGroViD).

{\small
\bibliographystyle{ieee}
\bibliography{egbib}

\begin{thebibliography}{10}\itemsep=-1pt

\bibitem{Ali2008}
S.~Ali and M.~Shah.
\newblock Floor fields for tracking in high density crowd scenes.
\newblock In {\em European Conference on Computer Vision}, pages 1--14, 2008.

\bibitem{Bailer2015}
C.~Bailer, B.~Taetz, and D.~Stricker.
\newblock Flow fields: Dense correspondence fields for highly accurate large
  displacement optical flow estimation.
\newblock In {\em International Conference on Computer Vision International
  Conference on Computer Vision}, 2015.

\bibitem{Middlebury}
S.~Baker, S.~Roth, D.~Scharstein, M.~J. Black, J.~P. Lewis, and R.~Szeliski.
\newblock A database and evaluation methodology for optical flow.
\newblock In {\em International Conference on Computer Vision}, pages 1--8,
  2007.

\bibitem{Barron94}
J.~L. Barron, D.~J. Fleet, and S.~S. Beauchemin.
\newblock Performance of optical flow techniques.
\newblock {\em International Journal of Computer Vision}, 12:43--77, 1994.

\bibitem{Bochinski2016}
E.~Bochinski, V.~Eiselein, and T.~Sikora.
\newblock Training a convolutional neural network for multi-class object
  detection using solely virtual world data.
\newblock In {\em International Conference on Advanced Video and Signal-Based
  Surveillance}, pages 278--285, 2016.

\bibitem{Sintel}
D.~J. Butler, J.~Wulff, G.~B. Stanley, and M.~J. Black.
\newblock A naturalistic open source movie for optical flow evaluation.
\newblock In {\em European Conf. on Computer Vision}, pages 611--625, 2012.

\bibitem{Curtis2016}
S.~Curtis, A.~Best, and D.~Manocha.
\newblock Menge: A modular framework for simulating crowd movement.
\newblock {\em Collective Dynamics}, 1(0), 2016.

\bibitem{Dehghan2018}
A.~Dehghan and M.~Shah.
\newblock Binary quadratic programing for online tracking of hundreds of people
  in extremely crowded scenes.
\newblock {\em IEEE Transaction on Pattern Analysis and Machine Intelligence},
  40(3):568--581, 2018.

\bibitem{Dosovitskiy2015}
A.~Dosovitskiy, P.~Fischer, E.~Ilg, P.~H{\"a}usser, C.~Haz{\i}rba{\c{s}},
  V.~Golkov, P.~v.d. Smagt, D.~Cremers, and T.~Brox.
\newblock Flownet: Learning optical flow with convolutional networks.
\newblock In {\em International Conference on Computer Vision}, 2015.

\bibitem{Dupont2017}
C.~Dupont, L.~Tob\'{i}as, and B.~Luvison.
\newblock Crowd-11: A dataset for fine grained crowd behaviour analysis.
\newblock In {\em Conference on Computer Vision and Pattern Recognition
  Workshops}, pages 2184--2191, 2017.

\bibitem{Kitti2012}
A.~Geiger, P.~Lenz, and R.~Urtasun.
\newblock Are we ready for autonomous driving? the kitti vision benchmark
  suite.
\newblock In {\em Conference on Computer Vision and Pattern Recognition}, pages
  3354--3361, 2012.

\bibitem{Geistert2016}
J.~Geistert, T.~Senst, and T.~Sikora.
\newblock Robust local optical flow: Dense motion vector field interpolation.
\newblock In {\em Picture Coding Symposium}, pages 1--5, 2016.

\bibitem{LearningScene-specificPedestrianDetectors}
H.~Hattori, V.~N. Boddeti, K.~Kitani, and T.~Kanade.
\newblock Learning scene-specific pedestrian detectors without real data.
\newblock In {\em Conference on Computer Vision and Pattern Recognition}, pages
  3819--3827, 2015.

\bibitem{Hu2017}
Y.~Hu, Y.~Li, and R.~Song.
\newblock Robust interpolation of correspondences for large displacement
  optical flow.
\newblock In {\em Conference on Computer Vision and Pattern Recognition}, pages
  4791--4799, 2017.

\bibitem{Hu2016}
Y.~Hu, R.~Song, and Y.~Li.
\newblock Efficient coarse-to-fine patch match for large displacement optical
  flow.
\newblock In {\em Conference on Computer Vision and Pattern Recognition}, pages
  5704--5712, 2016.

\bibitem{Haroon2014}
H.~Idrees, N.~Warner, and M.~Shah.
\newblock Tracking in dense crowds using prominence and neighborhood motion
  concurrence.
\newblock {\em Image and Vision Computing}, 32(1):14--26, 2014.

\bibitem{Ilg2017}
E.~Ilg, N.~Mayer, T.~Saikia, M.~Keuper, A.~Dosovitskiy, and T.~Brox.
\newblock Flownet 2.0: Evolution of optical flow estimation with deep networks.
\newblock In {\em Conference on Computer Vision and Pattern Recognition}, 2017.

\bibitem{JacquesJunior2010}
J.~C.~S. {Jacques Junior}, R.~S. Musse, and J.~R. Cl{\'{a}}udia.
\newblock {Crowd analysis using computer vision techniques}.
\newblock {\em IEEE Signal Processing Magazine}, (September):66--77, 2010.

\bibitem{Jodoin2013}
P.~M. Jodoin, Y.~Benezeth, and Y.~Wang.
\newblock Meta-tracking for video scene understanding.
\newblock In {\em International Conference on Advanced Video and Signal Based
  Surveillance}, pages 1--6, 2013.

\bibitem{Kroeger2016}
T.~Kroeger, R.~Timofte, D.~Dai, and L.~J.~V. Gool.
\newblock Fast optical flow using dense inverse search.
\newblock In {\em European Conference on Computer Vision}, pages 471--488,
  2016.

\bibitem{Kuhn2012}
A.~Kuhn, T.~Senst, I.~Keller, T.~Sikora, and H.~Theisel.
\newblock {A Lagrangian Framework for Video Analytics}.
\newblock In {\em Workshop on Multimedia Signal Processing}, pages 387--392,
  2012.

\bibitem{Li2015}
T.~Li, H.~Chang, M.~Wang, B.~Ni, and R.~Hong.
\newblock {Crowded Scene Analysis : A Survey}.
\newblock {\em Transactions on Circuits and Systems for Video Technology},
  25(3):367--386, 2015.

\bibitem{Liu2008}
C.~Liu, W.~T. Freeman, E.~H. Adelson, and Y.~Weiss.
\newblock {Human-assisted motion annotation}.
\newblock In {\em Computer Vision and Pattern Recognition}, pages 1--8, 2008.

\bibitem{LearningAppearanceInVirtualScenarios}
J.~Mar\'{i}n, D.~V\'{a}zquez, D.~Ger\'{o}nimo, and A.~M. L\'{o}pez.
\newblock Learning appearance in virtual scenarios for pedestrian detection.
\newblock In {\em Computer Society Conference on Computer Vision and Pattern
  Recognition}, pages 137--144, 2010.

\bibitem{Mehran2010}
R.~Mehran, B.~E. Moore, and M.~Shah.
\newblock {A streakline representation of flow in crowded scenes}.
\newblock In {\em European Conference on Computer Vision}, pages 439--452,
  2010.

\bibitem{Kitti2015}
M.~Menze and A.~Geiger.
\newblock Object scene flow for autonomous vehicles.
\newblock In {\em Conference on Computer Vision and Pattern Recognition}, pages
  3061--3070, 2015.

\bibitem{Narain2009}
R.~Narain, A.~Golas, S.~Curtis, and M.~C. Lin.
\newblock Aggregate dynamics for dense crowd simulation.
\newblock {\em ACM Transaction on Graphics}, 28(5):122:1--122:8, Dec. 2009.

\bibitem{SurveillanceInVirtualReality}
F.~Z. Qureshi and D.~Terzopoulos.
\newblock Surveillance in virtual reality: System design and multi-camera
  control.
\newblock In {\em Conference on Computer Vision and Pattern Recognition}, pages
  1--8, 2007.

\bibitem{Robicquet2016}
A.~Robicquet, A.~Sadeghian, A.~Alahi, and S.~Savarese.
\newblock Learning social etiquette: Human trajectory prediction in crowded
  scenes.
\newblock In {\em European Conference on Computer Vision}, pages 549--565,
  2016.

\bibitem{Senst2017}
T.~Senst, V.~Eiselein, A.~Kuhn, and T.~Sikora.
\newblock Crowd violence detection using global motion-compensated lagrangian
  features and scale-sensitive video-level representation.
\newblock {\em IEEE Transactions on Information Forensics and Security},
  12(12):2945--2956, 2017.

\bibitem{Weinzaepfel2013}
P.~Weinzaepfel, J.~Revaud, Z.~Harchaoui, and C.~Schmid.
\newblock Deepflow: Large displacement optical flow with deep matching.
\newblock In {\em Intenational Conference on Computer Vision}, pages
  1385--1392, 2013.

\end{thebibliography}
}

\end{document}